\documentclass[letterpaper]{article}
\usepackage[hyperref]{acl2021}
\usepackage{times}
\usepackage{latexsym}
\usepackage{multirow}
\usepackage{CJKutf8}
\usepackage{multirow}
\usepackage{titlesec}
\usepackage{caption}
\usepackage{graphicx}
\usepackage{amsmath}
\usepackage{mathtools}
\usepackage{times}
\usepackage{latexsym}

\usepackage{microtype}

\aclfinalcopy 

\title{DCT: Dynamic Compressive Transformer for Modeling Unbounded Sequence}

\author{Kai-Po Chang \\
  National Tsing Hua University \\
  \texttt{irving870810@gapp.nthu.edu.tw} \\\And
  Wei-Yun Ma \\
  Academia Sinica \\
  \texttt{ma@iis.sinica.edu.tw} \\}

\date{}

\begin{document}
\maketitle
\begin{abstract}
In this paper, we propose Dynamic Compressive Transformer (DCT), a transformer-based framework for modeling the unbounded sequence. In contrast to the previous baselines which append every sentence representation to memory, conditionally selecting and appending them is a more reasonable solution to deal with unlimited long sequences. Our model uses a policy that determines whether the sequence should be kept in memory with a compressed state or discarded during the training process. With the benefits of retaining semantically meaningful sentence information in the memory system, our experiment results on Enwik8 benchmark show that DCT outperforms the previous state-of-the-art (SOTA) model.
\end{abstract}

\section{Introduction}

When studying a textbook with tens of thousands of words, humans used to selectively memorize the important relevant knowledge from what we have learned. With this habit, most of the trivial details of text are discarded by us automatically, and we only retain the most essential knowledge in our brain.

Just like humans, artificial neural network also has memory systems to record what it has learned before. Early models like Recurrent Neural Networks (RNN, \cite{hochreiter1997long}) such as Long Short-Term Memory (LSTM, \cite{rumelhart1986learning}) use various learning gates and cell states to emulate a memory \cite{staudemeyer2019understanding} to store information from earlier time steps. Although these models are good for processing sequential data, they are difficult to optimize due to the gradient vanishing and explosion problem on long-range context \cite{hochreiter2001gradient}. Furthermore, previous work has \cite{khandelwal2018sharp} found that LSTM could only take advantage of 200 prior tokens on average. 

Afterwards, the transformer-based language model has achieved a wide range of impressive success \cite{devlin2018bert} by increasing the context length on language modeling \cite{dai2019transformer, rae2019compressive}, machine translation \cite{vaswani2017attention} and a large number of natural language processing (NLP) benchmarks \cite{devlin2018bert, yang2019xlnet}. The self-attention mechanism \cite{vaswani2017attention} contributes to the success of transformer, while the huge amount of parameters for integrating the hidden states in each layer makes itself limited to a fixed context length. With the increasing length of context cost bringing rising computational cost, the model has its limitation about how much information it can store in memory, which arises the long-range text modeling issue. 

To address issues from modeling long-range text, a dizzying number of \emph{"X-former"} models have been proposed \cite{tay2020efficient}. The main ideas on these models are fixed patterns \cite{child2019generating, ho2019axial, qiu2019blockwise}, learnable patterns \cite{kitaev2020reformer, tay2020sparse, roy2021efficient}, low rank methods \cite{choromanski2020rethinking, wang2020linformer, katharopoulos2020transformers}, and recurrence \cite{dai2019transformer, rae2019compressive}. In addition, all these solutions are either making improvements around computational and memory complexity or caching past representations in memory. Instead, the dynamically selecting mechanism of the compression judger also prevents us from excessively cost on self-attention mechanism \cite{vaswani2017attention} by discarding semantically meaningless sentences, which is also an approach to reduce the computational and memory complexity. Although these models make the transformer more efficient, they still cannot deal with the infinite context.

In this paper, we propose DCT, a dynamic compressive transformer for unbounded sequence modeling. Our framework consists of two models, a compressive transformer \cite{rae2019compressive} and a deep reinforcement-learning (RL) based compression judger, which determines the past sentence representation should be either kept in the memory system or discarded. In DCT, the compression judger is in charge of selecting the semantically meaningful sentence representation and thus the transformer could optimize task loss with the maximum information. As opposed to transformerXL \cite{dai2019transformer} and compressive transformer \cite{rae2019compressive} which append all past sentence representations to their memory system, our memory system can be much helpful for optimizing task loss. Also, we conduct our experiment on a character-level language modeling benchmark Enwik8 taken from the Hutter Prize \cite{kriesi2012political}, to demonstrate that our framework applies to learning long-range sequence representation. In summary, our contributions can be summarized as follows:
\begin{itemize}
    \item We propose DCT, Dynamic Compressive Transformer, that focuses on modeling sequence in unlimited long-range sequence.
    \item We present a transformer-based framework with an RL method that allows models to learn a more informatic memory system. DCT would drop the memories which are considered less important.
    \item We show that our method is a selective memorization process instead of keeping the latest sentence representation. This approach is applicable in searching the critical features in the dataset, and it could contribute to other NLP task where long context comprehension is needed.
\end{itemize}
\section{Related Work}
Many researches \cite{tay2020efficient} have attempted to capture the unbounded long-range context problem in language modeling. In recent years, improving transformer-based language models with memory mechanisms and reducing the computational requirements of attention mechanism are both of the most popular directions in particular. Previous studies about self-attention computational optimization include the sparse attention mechanism \cite{child2019generating, beltagy2020longformer, tay2020sparse, roy2021efficient} and self-attention time and memory complexity reduction models \cite{kitaev2020reformer, choromanski2020rethinking, peng2021random}. Although these models and techniques can reduce the computational requirements and memory cost, all memories need to be kept during training. And their performances on Enwik8 dataset do not exceed transformerXL \cite{dai2019transformer}.

On the other hand, leveraging a side memory module that caching past representations \cite{dai2019transformer, rae2019compressive} has been regarded as the most potent approach to address the computational problem on the unbounded input sequences. Since \citeauthor{dai2019transformer} (\citeyear{dai2019transformer}) proposed the TransformerXL that uses a novel relative position embedding and a segment-level recurrence mechanism to keep past information in memory, they reached SOTA results on multiple benchmarks including WikiText-103 \cite{merity2016pointer} and Enwik8 \cite{hutter2012human}. Following this, \citeauthor{rae2019compressive} (\citeyear{rae2019compressive}) proposed compressive transformer which incorporated the ideas of transformerXL \cite{rae2019compressive} to preserve prior sentences and furthermore improves its memory system through both short-term granular memory and longer-term coarse compressed memory. One problem of these approaches is they would eventually drop the oldest memories regardless of the importance of the memory in the current circumstance whenever the memory system is full. In comparison, our framework has a procedure to decide the oldest memory when the oldest sentence is beneficial to task performance. Recently, the infinity transformer \cite{martins2021infty} leverages continuous attention \cite{martins2020sparse} to read unbounded context. Although it is demonstrated its ability on their synthetic sorting task, it (24.41 ppl) only gains slight improvement on Wikitext-103 \cite{merity2016pointer} than the previous SOTA model (24.22 ppl) on their experiment. Finally, inspired by compressive transformer \cite{rae2019compressive}, DCT uses it as part of our framework to preserve both the short-term and long-term information. 


\section{Models}
The task we cope with in this paper is language modeling whose metrics for evaluating are Perplexity (PPL) and Bits-Per-Character (BPC). Given a tokenized sequence $\mathbf{w} = \{w_0, w_1, ..., w_T\}$, and the perplexity of $\mathbf{w}$ is 
\begin{equation} 
\operatorname{PPL}(\mathbf{w})=\exp \left\{-\frac{1}{T} \sum_{i}^{T} \log p_{\theta}\left(w_{i} \mid  w_{<i}\right)\right\} \label{eq:one}
\end{equation}

where $\log p_{\theta}\left(w_{i} \mid w_{<i}\right)$ is the log-likelihood of the i-th token conditioned on the preceding tokens $w_{<i}$ \cite{wolf-etal-2020-transformers}, while BPC is defined as the average cross-entropy (used with log base 2) \cite{graves2013generating}. A language model is a probability distribution over entire context, so these two metrics could be considered as an evaluation of the model's ability to predict next specified token according to the conditional probability (like in Eq.~\ref{eq:one} ).  
    
\begin{figure*}
    \centering
    \includegraphics[width=\textwidth]{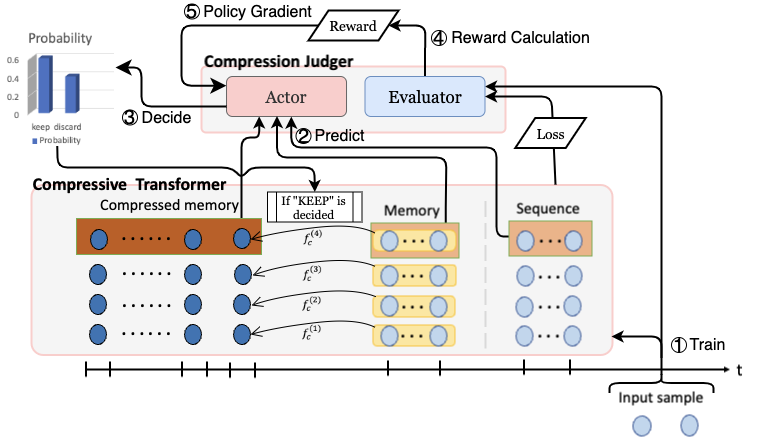}
    \caption{DCT: Dynamic compressive transformer. The process in this diagram is executed per step. The above model has four layers with a sequence length $n_{s} = 128$, a memory size $n_{m}=128$, a compressed memory size $n_{cm}=256$, and a compression ratio of $c$. First, the yellow highlighted memories will be compressed through a compression function $f_{c}$ in each layer. Second, the compression judger would decide whether the entire layers should be kept in compressed memory or be discarded.}
    \label{fig:one}
\end{figure*}

We propose DCT, a transformer-based framework that determines whether to keep or discard the compacted past activations into a compressed memory. As illustrated in Fig.~\ref{fig:one}, DCT is composed of the compression judger and compressive transformer \cite{rae2019compressive}. When compressive transformer \cite{rae2019compressive} eliminates these oldest past activations from memory, the compression judger will make a binary decision on each input evicted memory to decide whether they should be appended to compressed memory or be discarded. 

\subsection{Compressive Tranasformer}
Compressive transformer \cite{rae2019compressive} is a long-range sequence model which compresses and append all past activations into compressed memory. They applies a \emph{compression operation} $f_{c}: \mathbf{R}^{n_{s} \times d} \rightarrow \mathbf{R}^{\left[\frac{n_{s}}{c}\right\rfloor \times d}$ which is an 1D convolution layer for mapping the $n_s$ oldest memories to $\left\lfloor\frac{n_{s}}{c}\right\rfloor$ compressed memories \cite{rae2019compressive}. In addition, those compacted momories will be stored in secondary FIFO compressed memory. However, it still needs to discard these compressed memories when the compressed memory is full.

Therefore, we modify and improve its original operation to be \emph{selectively compression operation}. Besides compressing and appending all the eliminated memories to secondary compressed memory, we also apply a compression judger that determines whether the compressed memories are beneficial for the transformer to minimize the task loss depends on the content of compressed memory, memory, and sequence. In other words, the compression judger could select and append the meaningful compacted sentence representation to compressed memory, instead of preserving all text including meaningless sentences.

\subsection{Compression Judger}
We interpret the problem of how to identify the semantically important sentences based on reinforcement learning. The state \emph{s} is concatenated tensors composed of the last layer of compressed memory, memory, and sequence [($n_{cm}$ $+$ $n_{m}$ $+$ $n_{s}$) $\times$ dimension]. The action space \emph{a} is discrete in our framework and given as \{0 (\emph{Discard}), 1 (\emph{Keep})\}, which means to append or discard the compacted memories, respectively. Depends on state, the actor, parameterized by $\theta$, follows a policy $p_{\theta}(a \mid s)$ to sample an action from action set. Although the policy $p_{\theta}(a \mid s)$ is stochastic, the state transition is deterministic after an action has been choosen, i.e. $\delta_{s, s^{\prime}}^{a} = 1$. That is, the problem could be formulated as Markov Decision Process (MDP), which is well aligned with RL problem \cite{barto2017some}. 

Therefore, we propose the RL-based compression judger which is composed of two components, actor and evaluator. The actor makes a binary decision for determining whether to append the evicted memories to compressed memory, and the evaluator is a freezing pre-trained language model to estimate the perplexity of the input sentence. The details are described as following:

\subsubsection{Actor}
The main architecture of the actor is a BiLSTM encoder \cite{schuster1997bidirectional}. To model the actor as a policy of selecting the evicted old memory more accurately, we treat the actor as an agent and adopt the policy gradient method \cite{sutton2000policy} to optimize the actor by maximizing the objective function $J(\theta)$:
\begin{equation}
J(\theta)=E_{\tau \sim p_{\theta}(\tau)}\left[R(\tau)\right]=\sum_{\tau} R(\tau) p_{\theta}(\tau)\label{eq:two}
\end{equation}
where $\tau$ is given as $\{s_1, a_1, r_1,..., s_n, a_n, r_n, ..., s_N, $ $a_N, r_N\}$ which is known as trajectory or episode, $\theta$ is the parameter of the actor, $p_{\theta}(\tau)$ represents the softmax probability of the sampled determination, and $R(\tau)$ will be discussed later. The trajectory begins when both the compressed memory and memory are full so that the tensor shape of state would be fixed. In timestep $\emph{n}$, the actor takes specific action $a_n$ (keep or discard) according to current $s_n$ (the last layer of memory system and input sequence), and then obtains the corresponding reward $r_n$. 

The next question is how to estimate the total reward of trajectory $R(\tau)$. In this paper, we consider the perplexity from compressive transformer as the reward. However, the property of perplexity is the lower the better, which doesn't conform to the standard reward definition in RL problems like OpenAIGym \cite{brockman2016openai}, so we apply the perplexity (ppl) with the following function $f(.)$:
\begin{equation}
f(ppl)= m \cdot a^{ppl} \label{eq:three}
\end{equation}
where the base nubmer $a$ and slope $m$ are hyperparameters, and $a$ must be between $0$ and $1$. With this function, we could make the reward and the transformed perplexity with a positive correlated relation. Thus, we would make the reward the higher the better by redesigning it as follows if we sample a trajectory $\tau$ with $N$ state-action pairs.
\begin{equation}
R(\tau) =\frac{1}{N} \sum_{t=1}^{N} \left\{r_{t} = f\left(ppl_{t}\right)  \label{eq:four} \right\}
\end{equation}

\subsubsection{Evaluator}
The evaluator is an arbitrary freezing pre-trained language model. Due to the high variance in evaluating the perplexity between different sequences when using REINFORCE algorithm \cite{williams1992simple}, we need to set a baseline $b$ to mitigate this problem to maximize the reward $R(\tau)$.

Once a trajectory has been processed by the DCT, we are ready to update the actor. In fact, each state-action pair in a trajectory is collected through one batch, and the details will be discussed in Section~\ref{sec:three_three}. Following \citeauthor{sutton2000policy} (\citeyear{sutton2000policy}), to maximize the total reward of trajectory $R(\tau)$, the gradient of the objective can be derived as:
\begin{equation}
\begin{multlined}
\nabla_{\theta} J(\theta) = E_{\tau \sim p_{\theta}(\tau)}\Big[ (R(\tau)-b+\alpha \cdot S[p_{\theta}]) \\ 
\nabla \log p_{\theta}\left(s_{t}, a_{t}\right)\Big] \label{eq:five}
\end{multlined}
\end{equation}
where $b$ is also an averaged transformed perplexity score from a freezing pre-trained language model in one trajectory like $R\left(\tau\right)$, $\alpha$ is a coffeicent and $S[p_{\theta}]$ denotes an entropy of the action probability from actor. 

Since it is not possible to enumerate all possibilities of the real samples, we apply the REINFORCE algorithm \cite{williams1992simple} which relies on Monte-Carlo based policy gradient methods; it works because the expectation of the sample gradient as follows is approximated to the actual gradient in Eq.~\ref{eq:five}:
\begin{equation}
\begin{multlined}
\nabla_{\theta} J(\theta) \approx \frac{1}{N} \sum_{t=1}^{N} \Big[(r_{t}-b+\alpha \cdot S[p_{\theta}]) \\ \nabla \log p_{\theta}\left(s_{t}, a_{t}\right)\Big] \label{eq:six}
\end{multlined}
\end{equation} 

The above formula is the process during optimization. We use the gradient ascent to find the best actor $\theta$ that produces the highest reward. Then we update the actor's parameters with $\eta$ denotes the learning rate as:
\begin{equation}
\theta \leftarrow \theta+\eta \nabla_{\theta} J(\theta) \label{eq:seven}
\end{equation}

\subsection{Training Description and Details} \label{sec:three_three}
To train the proposed DCT, we need to optimize the two components, compressive transformer \cite{rae2019compressive} and compression judger, simultaneously. We propose the following training procedures during each batch. In one batch, we divide it into mini-batches and go through them one by one. In this way, we can append past hidden states in the unit of minibatch, reduce the computational cost, and synchronize the updating frequency of compressive transformer \cite{rae2019compressive} and compression judger (see Figure ~\ref{fig:one}).
\begin{enumerate}
    \item \textbf{DCT training :}  
    We first pretrain the compressive transformer \cite{rae2019compressive} for nearly one epoch, and then combine it with the training of the compression judger. At beginning of each step, we split the batch of samples into several smaller mini-batches of samples; then train both of the Compressive transformer \footnote{We use and modify the author's implementation available at https://github.com/lucidrains/compressive-transformer-pytorch} \cite{rae2019compressive} and freezing evaluator in compression judger with the same mini-batch as input sample at the same time.
    \item \textbf{Compression judger prediction :} 
    The policy (agent behavior strategy) $p_{\theta}\left(a \mid s\right)$ is predicted by the actor in compression judger: $$p_{\theta}\left(a \mid s\right) = \mbox{softmax}( \mbox{FC}( \mbox{BiLSTM}(s))) \label{eq:eight}$$ where $s$ is the concatenated sentence representations of compressed memory, memory, and sequence. The reason to use BiLSTM encoder \cite{schuster1997bidirectional} is that \citeauthor{parisotto2020stabilizing} (\citeyear{parisotto2020stabilizing}) demonstrated the effectiveness of LSTM \cite{rumelhart1986learning} agent in RL problem outperform transformer-XL \cite{dai2019transformer} agent. 
    
    In compression judger, the actor is in charge of predicting an action probability distribution $p_{\theta}\left(a \mid s\right)$ after the compressed memory and memory are filled with the pasted hidden states. And we wouldn't discard the past activations unless the memory system is full.
    \item \textbf{Compression judger decision :}
    Depends on the policy $p_{\theta}\left(a \mid s\right)$ of this binary decision in Eq.~\ref{eq:eight}, the compression judger sample an action from the action probability to determine evicted memories would be discarded or appended to compressed memory in a compressed state.
    \item \textbf{Reward calculation :}
    After sequentially passing each mini-batch to both of the compressive transformer \cite{rae2019compressive} and evaluator, we get two lists that contain the perplexity scores from each individual models. In order to decrease the high variance of sentences, we do more than using per scores from transformer as reward. Instead, we decrease those by a baseline $b$, the perplexity scores from evaluator, to make the final return values.
    $$b =\mbox{cross-entropy}(\mathrm{logits}, \mathrm{target})$$
    $$\small{\mathrm{logits} = \mbox{FC}(\mbox{Freezing-GPT2}(\mathrm{input}\textendash \mathrm{smaples}))}$$
    where input-samples are the mini-batch of samples, and target is the shifted mini-batch with one token difference. In our implementation, the freezing pre-trained language model is HuggingFace's pre-trained OpenAI-GPT2\footnote{https://github.com/huggingface/transformers} \cite{radford2019language}.
    
    \item \textbf{Policy gradient :}
    A trajectory begins after the memory system arefull, and it will be finished when one step is finished. The actor will optimize its parameters $\theta$ through policy gradient \cite{sutton2000policy} to maximize the reward $r_{t}$ in Eq.~\ref{eq:six}.
    
\end{enumerate}

\section{Experiments}
\subsection{Dataset}
Our method is evaluated on Enwik8 dataset, a standard character-level language modeling benchmark taken from Hutter Prize \cite{hutter2012human}, which contains 100M bytes of unprocessed Wikipedia text. The first 90MB is the training dataset, the following 5MB is the validation dataset, and the last 5MB is the testing dataset.

\subsection{Implementation Settings and Details}
We set the batch size to 32 and sequence length to 128. For the memory system, the compressed memory length is 64, memory length and sequence length are 128 with a compression ratio of 4. As for the learning rate, we use fixed $1\times 10^{-4}$ when pretraining compressive transformer, and then co-training with compression judger with fixed $1\times 10^{-5}$ on both models. Also, we train our baseline model, compressive transformer \cite{rae2019compressive}, with the same hyperparameters.



\subsection{Results}
The testing results of DCT are shown in Table~\ref{table:eval}. As we can see, Ours-8L DCT outperforms other models in BPC on Wiki8 dataset \cite{hutter2012human}. The results show that increasing the reading length by the revised memory mechanism leads to better perplexity.

\begin{table}[h]
\setlength{\arrayrulewidth}{0.5mm}
\renewcommand{\arraystretch}{1.2}
\begin{center}
\begin{tabular}{p{6cm}|c} 
 \hline
 Models & BPC \\
 \hline \hline
 7L LSTM \cite{graves2013generating} & 1.67 \\
 12L Transformer \cite{al2019character} & 1.11\\
 12L Transformer-XL \cite{dai2019transformer} & 1.06\\
 24L Transformer-XL \cite{dai2019transformer} & 0.99\\
 24L Compressive Transformer \cite{rae2019compressive} & 0.97\\
 \hline
 Ours-8L Compressive Transformer & 0.978\\
 Ours-8L DCT & \bf{0.963}\\
 \hline
\end{tabular}
\end{center}
\caption{State-of-the-art testing results on Enwik8.}
\label{table:eval}
\end{table} 

We observe that DCT is able to outperform the compressive transformer \cite{rae2019compressive} in EnWik8. However, we only trained our model for at most 10 days on a single GPU in small sequence length and small-batch settings. It would be advantageous to train our DCT with the large batch size and sequence length for its memory selecting mechanism. 

\subsection{Analysis}
\begin{figure}[h]
    \centering
    \includegraphics[width=\linewidth]{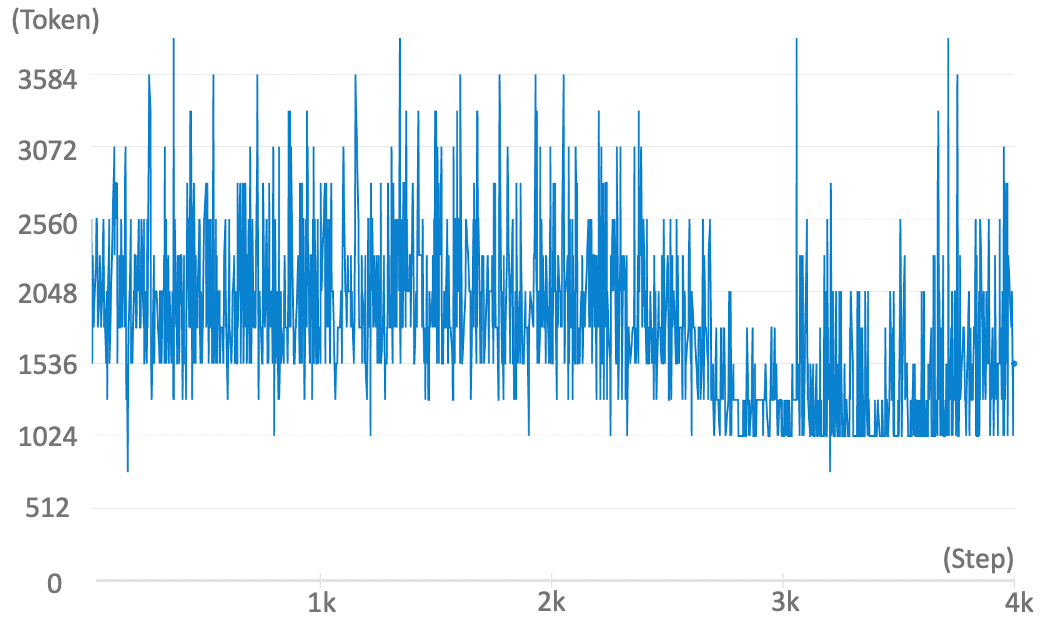}
    \caption{The reading distance for each batch.}
    \label{fig:longest}
\end{figure}

To know whether the compression judger contributes to the reduction in complexity, we show the Figure.~\ref{fig:longest} which is the results of the longest length read by DCT for each batch. The definition of longest length is the most distant tokens of all hidden vectors corresponding to the corpus in the memory system. The shorter the reading distance of the model in a certain range, the more the model chooses to keep the evicted memories in compressed memory. In Figure.~\ref{fig:longest}, we observe that the reading distance of DCT is shorter than average between the 2500-th and 3500-th step, which means the compression judger tends to compress the evicted memories in this range instead of discarding them. This represents the compression judger regards this range of text as more highly significant information for optimizing the task loss, which demonstrates the ability of DCT to identify the critical information in a corpus.

Inspired by SeqGAN \cite{yu2017seqgan}, we first pretrain the compressive transformer to generate the more semantically sentence embedding, and then co-train both of the compressive transformer and compression agent to reduce the BPC. Besides, the entropy bonus $S[p_{\theta}]$ could also help training because it avoids the overly lopsided view of compression judger that everything should be discarded or kept forever. These two training skills help us construct a better language modeling method, the RL-based co-training framework of the compressive transformer and compression judger.

\section{Conclusion}
In this paper, we propose the DCT, a framework extended from compressive transformer. By making use of an RL-based compression judger which is composed of actor and evaluator, the DCT could capture the further long context and even the unbounded sequence with computation budget. Instead of keeping all information in the memory system, the actor in DCT determines whether to append the past activations or discard them based on the current memory system and input samples. Namely, the actor judge whether the past activations are beneficial to our language modeling task in the attention calculation. Our experiments show significant improvement in BPC compared with previous SOTA models on Enwik8 dataset. 

\section{Future Work}
Inspired by SimCSE \cite{gao2021simcse}, our next step would be pretraining compressive transformer in DCT on Natural language inference (NLI) dataset \cite{bowman2015large} with contrastive learning objective. Thus, the actor in DCT could base on the more semantically meaningful sentence representations to judge the importance of the past activation to the language modeling task. Another future work is to apply our approach to other NLP tasks such as multiple-document summarization or medical information problem tasks based on long DNA sequences.

\bibliographystyle{acl_natbib}
\bibliography{anthology,acl2021}


\end{document}